%% file: main.tex
\documentclass[10pt,twocolumn,letterpaper]{article}

\usepackage{cvpr}              
\usepackage{booktabs}
\usepackage{multirow}
\usepackage{amssymb}
\usepackage{threeparttable}


\definecolor{cvprblue}{rgb}{0.21,0.49,0.74}
\usepackage[pagebackref,breaklinks,colorlinks,allcolors=cvprblue]{hyperref}
\usepackage{cuted}
\usepackage{capt-of}
\usepackage{placeins}
\usepackage{stfloats}

\def\confName{CVPR}
\def\confYear{2027}

\title{%
    \makebox[\textwidth][c]{%
    \includegraphics[height=1.5cm]{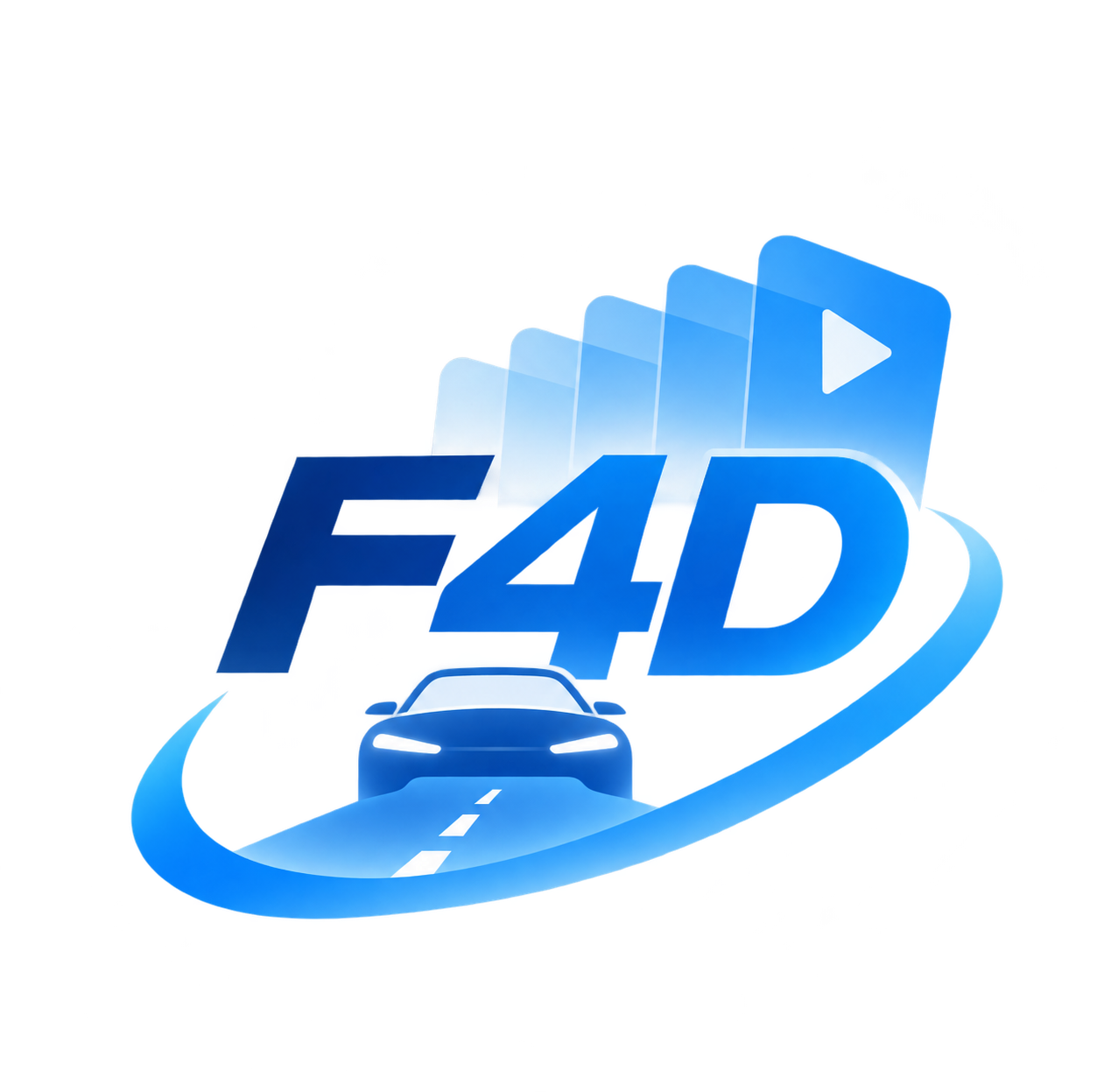}%
        \hspace{0.35cm}%
        {\fontsize{14}{17}\selectfont
        \shortstack[c]{%
            \textbf{FreeForward4D}: \textsc{Annotation-Free Feed-Forward}\\[2pt]
            \textsc{4D Reconstruction of Driving Scenes}
        }}%
    }%
}

\author{
Luosong Guo$^{1,2,*}$,
Xiang Yang$^{1,2*}$,
Chaoyang Li$^{1,2}$,
Ao Gao$^{1}$,
Jiangming Shi$^{1}$, \\
Yanyun Qu$^{3}$,
Zhizhong Zhang$^{1}$,
Jingyu Gong$^{1}$,
Yuan Xie$^{1,2,\dagger}$ \\[0.4em]
{
$^{1}$East China Normal University \quad
$^{2}$Shanghai Innovation Institute} \\
{
$^{3}$Xiamen University} \\[0.3em]
{
$^{*}$Equal contribution. \quad
$^{\dagger}$Corresponding author.}
}

\begin{document}
\maketitle
\input{sec/teaser}  
\input{sec/0_abstract}

\input{sec/1_intro}
\input{sec/relatework}

\begin{figure*}[!t]
\centering
\includegraphics[width=\textwidth]{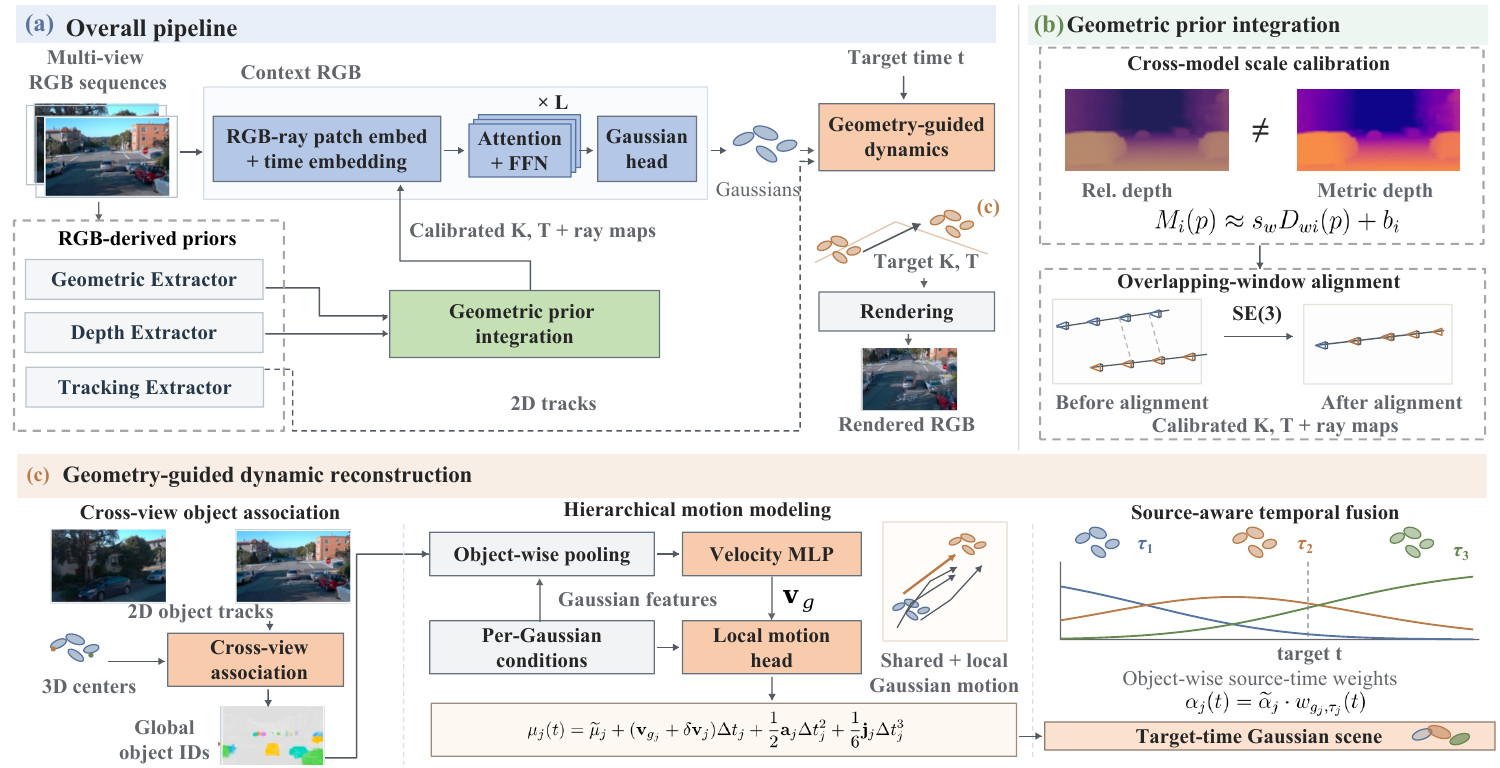}
\caption{\textbf{FreeForward4D pipeline.}
(a) Frozen visual priors provide windowed cameras, pseudo-depth, and 2D object tracks. (b) Geometric prior integration calibrates the windows against monocular metric depth and aligns them into a single metric frame (before vs.\ after alignment). (c) This geometry enables cross-view object
association, object motion is modeled by a shared velocity and local
Gaussian trajectories, and temporal weights aggregate Gaussians from
different context times.
The dashed line carries 2D tracks.}
\label{fig:ff_pipeline}
\end{figure*}

\input{sec/method}

\input{sec/experiments}

\input{sec/Conclusion}

\section*{LLM Usage Statement}
We used publicly available LLM assistants (ChatGPT/Codex, Gemini, and Kimi). LLM-based coding agents were used extensively, under the authors' direction, to implement and debug code for data preprocessing, training, and evaluation, and to run experiments. LLM assistants were also used to edit the manuscript (wording and grammar), to check consistency among text, tables, figures, and references, and to adjust figure labels and layout. The research ideas and method design originate from the authors. The authors verified all code, results, and claims, and take full responsibility for the content of this paper.

{
    \small
    \bibliographystyle{ieeenat_fullname}
    \bibliography{main,references}
}

\input{sec/supplement}
\end{document}

%% file: sec/teaser.tex
\begin{strip}
    \centering
    \includegraphics[width=\textwidth]{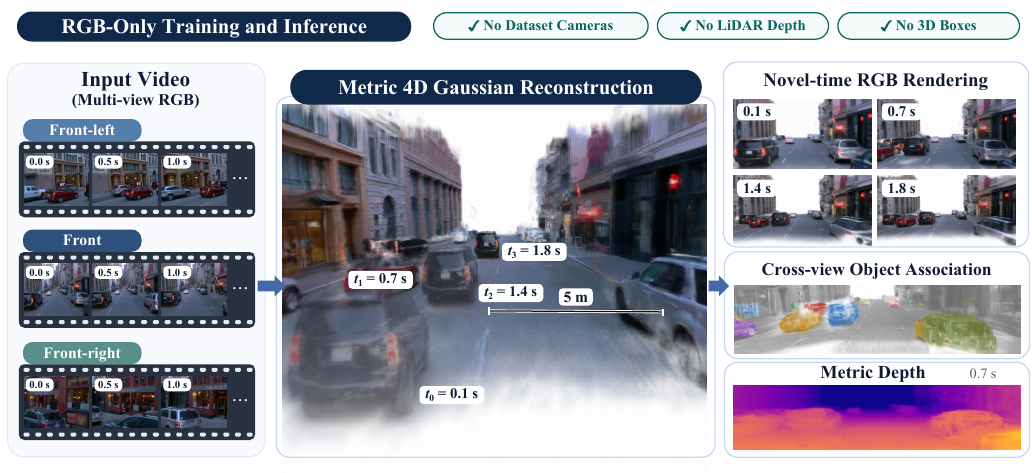}
    \captionof{figure}{\textbf{Annotation-free feed-forward reconstruction of dynamic driving scenes}. Given only multi-view RGB videos, FreeForward4D reconciles pretrained geometric priors into a single metric frame and reconstructs metric-scale 4D Gaussians without dataset-provided cameras, LiDAR depth, or 3D box annotations during training or inference. The reconstructed scene captures object dynamics and supports novel-time rendering at interpolated and extrapolated timestamps. We show a vehicle at $t_0$--$t_3$ (center), synchronized RGB renderings, cross-view object associations, and metric depth (right).}
    \label{fig:ff_teaser}
\end{strip}

%% file: sec/0_abstract.tex
\begin{abstract}
Feed-forward 4D reconstruction enables efficient creation of renderable driving scenes for autonomous driving simulation. However, most existing methods rely heavily on dataset-provided cameras, LiDAR depth, or 3D box annotations, limiting their applicability to unannotated driving videos. Pretrained visual models can predict cameras, depth, and object tracks from RGB alone, but their geometry is mutually inconsistent: multi-view predictions come in short windows with independent scales and frames, while monocular depth is metric only per image. Our key insight is that, once reconciled into a single metric frame, these priors can replace dataset annotations for feed-forward 4D reconstruction. We present \textbf{FreeForward4D}, which calibrates overlapping prediction windows against monocular metric depth and registers them into one frame across time. On this geometry, 2D tracks are associated across cameras in 3D, and objects move through hierarchical motion that combines a shared object velocity with local Gaussian corrections, so that both training and inference use only RGB videos. On Waymo, FreeForward4D achieves 27.66 dB PSNR and 0.836 SSIM while recovering metric-scale cameras and depth without scale alignment. It surpasses STORM, SLARM-F, UFO, DGGT, and StreetForward, which rely on dataset annotations, in full-image PSNR and attains the best RGB fidelity on 8-second windows. RGB-only fine-tuning further adapts the framework to nuScenes and Argoverse 2. Our code and models are available at: \url{https://github.com/Max-luo-song/FreeForward4D}.
\end{abstract}

%% file: sec/1_intro.tex
\section{Introduction}
\label{sec:intro}

Reconstructing renderable dynamic driving scenes from real-world videos is essential for scalable sensor simulation and closed-loop evaluation in autonomous driving. Existing per-scene methods achieve high-fidelity reconstruction by optimizing a dedicated representation for each driving sequence~\cite{ff_streetgaussians,ff_omnire,ff_emernerf}. However, their costly per-scene optimization, often requiring minutes to hours, makes scaling to large collections of driving videos impractical. This computational bottleneck has motivated feed-forward approaches, which learn transferable reconstruction priors across scenes and directly predict renderable 4D representations without scene-specific optimization~\cite{ff_storm,ff_drivingrecon,ff_ufo,ff_slarm}.

Despite these advances in efficiency, existing feed-forward methods still rely heavily on dataset annotations that are costly to acquire at scale. They typically require dataset cameras (intrinsics and poses) to establish scene geometry, LiDAR depth for geometric supervision, and often 3D box annotations or the object trajectories derived from them for dynamic modeling. Recent pose-free methods~\cite{ff_dggt,ff_streetforward} eliminate the need for camera poses at inference, but still rely on dataset annotations during training. Moreover, pose-free reconstruction is not guaranteed to be metric; DGGT, for instance, aligns its depth to LiDAR before evaluation~\cite{ff_dggt}. Consequently, these dependencies limit the applicability of feed-forward reconstruction to unannotated driving videos, where neither dataset annotations nor metric-scale references are available. This raises a fundamental question: \emph{Can feed-forward 4D reconstruction be learned from RGB videos alone, without dataset-provided cameras, LiDAR depth, or 3D box annotations during training or inference, while recovering metric-scale geometry and coherent scene dynamics?}

Pretrained visual foundation models now estimate cameras and depth from RGB images~\cite{ff_vggt,ff_omega,ff_moge2} and provide object-level observations without 3D box annotations~\cite{ff_groundingdino,ff_sam2}. Their geometry, however, is not directly usable for reconstruction, because it is inconsistent at three levels: in \emph{scale}, multi-view models predict relative depth within each temporal window while monocular models predict metric depth per image; in \emph{coordinate frame}, each window has its own reference; and in \emph{image grid}, predicted intrinsics refer to each model's input resolution rather than the reconstruction grid. Image-derived object tracks additionally lack cross-camera correspondence and 3D motion, while each timestamp observes only part of a moving object's surface. Combined naively, these priors yield cameras, depth targets, and tracks that disagree with one another.

Our key insight is that these priors, once reconciled into a single metric frame, can replace dataset cameras, LiDAR depth, and 3D boxes for both reconstruction and dynamics. Building on this insight, we present \textbf{FreeForward4D}, an annotation-free feed-forward framework for dynamic driving scene reconstruction using only RGB-derived conditions (Fig.~\ref{fig:ff_teaser}). First, metric-consistent geometric prior integration resolves all three inconsistencies: it calibrates the scale of each overlapping window against monocular metric depth, registers the windows into one scene frame, and maps the predicted intrinsics onto the reconstruction grid, yielding metric cameras and depth without ground-truth scale alignment. Second, geometry-guided dynamic reconstruction exploits this common frame to associate 2D tracks across cameras in 3D and to move each object with a shared velocity refined by local Gaussian trajectories. Source-aware temporal fusion further integrates motion-compensated observations across timestamps for novel-time rendering.

Extensive experiments on Waymo Open~\cite{ff_waymo}, nuScenes~\cite{ff_nuscenes}, and Argoverse2~\cite{ff_argoverse2} demonstrate the effectiveness of FreeForward4D. Pose-conditioned STORM, SLARM-F, and UFO require dataset cameras as input, and pose-free methods require dataset annotations for training; without either, FreeForward4D achieves higher full-image PSNR than these methods, with the highest RGB fidelity over extended 8-second windows. On Waymo, it further recovers metric-scale cameras without any scale fitting, and RGB-only fine-tuning suffices to adapt it to new datasets.

Our contributions are summarized as follows:
\begin{itemize}
    \item We present \textbf{FreeForward4D}, a feed-forward 4D reconstruction framework trained and run on RGB driving videos alone, and show that pretrained priors, once reconciled into a single metric frame, can replace dataset-provided cameras, LiDAR depth, and 3D box annotations.

    \item We propose metric-consistent geometric prior integration, which resolves inconsistencies in scale, coordinate frame, and image grid by calibrating overlapping windows against monocular metric depth, registering them into one metric frame, and aligning camera rays with the reconstruction grid, yielding metric cameras and depth without dataset cameras, LiDAR depth, or ground-truth scale alignment.

    \item We propose geometry-guided dynamic reconstruction, which associates per-camera 2D tracks across cameras through their 3D centers in this metric frame rather than 3D box annotations, couples them through a shared object velocity refined by local higher-order Gaussian motion, and fuses context times with per-object normalized weights.

    \item Experiments on Waymo Open, nuScenes, and Argoverse2 show that, without any dataset annotations, FreeForward4D surpasses STORM, SLARM-F, UFO, DGGT, and StreetForward in full-image PSNR, attaining the highest RGB fidelity on Waymo 8-second windows.

\end{itemize}

%% file: sec/relatework.tex
\section{Related Work}
\textbf{Per-scene driving scene reconstruction.} Per-scene driving reconstruction has been extensively explored using neural radiance fields~\cite{ff_nerf} and, more recently, 3D Gaussian Splatting~\cite{ff_gaussians}. These representations have also been extended to autonomous driving sensor simulation~\cite{ff_neurad,ff_splatad}, typically requiring scene-specific optimization for each recorded sequence. Existing methods model scene dynamics through two main paradigms. Object-centric approaches explicitly decompose driving scenes into static backgrounds and individual moving actors, whose motions are modeled using object-level transformations~\cite{ff_nsg,ff_mars,ff_unisim,ff_drivinggaussian,ff_streetgaussians,ff_hugs,ff_omnire}. Such methods typically rely on 3D bounding boxes or tracked object poses to initialize dynamic reconstruction. Scene-level approaches, in contrast, capture dynamics through spatiotemporal fields, scene flow, or time-varying Gaussian attributes, without requiring predefined object trajectories~\cite{ff_suds,ff_emernerf,ff_pvg,ff_s3gaussian,ff_desiregs}. 

Despite their different motion representations, both paradigms require computationally expensive scene-specific optimization, often taking hours per sequence and limiting their scalability to large-scale driving datasets.

\textbf{Feed-forward driving scene reconstruction.}
Feed-forward methods replace per-scene optimization with networks that directly predict radiance fields or 3D Gaussians from images~\cite{ff_pixelnerf,ff_ibrnet,ff_pixelsplat,ff_mvsplat}. In driving scenarios, sparse surround-view observations and complex scene dynamics pose challenges for both geometric reconstruction and temporal consistency. Existing methods address sparse-view geometry through depth estimation and multi-view feature aggregation~\cite{ff_drivingforward,ff_evolsplat,ff_ggs}, while dynamic reconstruction incorporates motion-aware aggregation, static--dynamic decomposition, and time-dependent Gaussian representations~\cite{ff_drivingrecon,ff_storm,ff_slarm}. Other approaches explore recurrent scene updates, object-centric representations, and future-time synthesis to improve temporal modeling and rendering flexibility~\cite{ff_ufo,ff_pointforward,ff_envision4d}. 
Recent pose-free methods jointly estimate cameras and dynamic representations from unposed images~\cite{ff_dggt,ff_streetforward}, but are still trained with dataset annotations. Flux4D~\cite{ff_flux4d} avoids 3D box annotations but requires dataset cameras and LiDAR as input. 

In contrast, FreeForward4D enables feed-forward 4D reconstruction from RGB driving videos alone, without requiring dataset-provided cameras, LiDAR depth, or 3D box annotations during training or inference.

\textbf{Foundation models for geometry and object priors.}
Large-scale visual pretraining has enabled transferable representations for diverse geometric and semantic tasks~\cite{ff_dinov2}. For geometric reconstruction, monocular models provide dense depth and metric geometry estimates~\cite{ff_depthanything2,ff_metric3dv2,ff_moge,ff_moge2}, while multi-view models recover cross-view correspondences and scene structure from uncalibrated images~\cite{ff_dust3r,ff_mast3r,ff_vggt,ff_pi3}. Recent advances in model scaling, recurrent inference, and chunk-wise alignment of long sequences have further improved multi-view geometric reconstruction~\cite{ff_omega,ff_cut3r,ff_vggtlong}. Combining monocular metric depth with multi-view geometry has been explored to recover the metric scale of a single reconstruction~\cite{chen2025spatial}; FreeForward4D builds on such calibration to reconcile scale, coordinate frame, and image grid across overlapping temporal windows, yielding a single metric frame that conditions feed-forward reconstruction. Beyond geometry, open-vocabulary detection and promptable segmentation provide object-level cues that can be propagated across video frames~\cite{ff_groundingdino,ff_sam,ff_sam2}.

%% file: sec/method.tex
\section{Method}
\label{sec:method}
\paragraph{Problem formulation.}
We study offline 4D reconstruction of a multi-view driving video $\mathcal I=\{I_t^v\}$, indexed by time $t$ and camera $v$, without dataset-provided cameras, LiDAR depth, or 3D box annotations in training or inference, which we refer to as annotation-free. Frozen visual priors process the full video to estimate camera intrinsics $\widehat K_t^v$, metric camera-to-world poses $\widehat T_t^v$, and 2D object tracks, and provide monocular pseudo-depth for scale calibration and supervision. A feed-forward reconstructor then predicts dynamic Gaussians and their motion from a sparse set of context frames $\mathcal C\subset\mathcal I$ and their derived conditions, which can be rendered at arbitrary target cameras and times. Given the precomputed priors, reconstruction and rendering run in real time (Sec.~\ref{sec:ff_experiments}).

\paragraph{Overview.}
As illustrated in Fig.~\ref{fig:ff_pipeline}, FreeForward4D consists of two components: metric-consistent geometric prior integration (Sec.~\ref{sec:ff_geometry}), which unifies heterogeneous visual predictions into globally consistent metric geometry, and geometry-guided dynamic reconstruction (Sec.~\ref{sec:ff_hierarchical_dynamics}), which establishes cross-view object associations and models coherent dynamics through hierarchical motion modeling and source-aware temporal fusion.

\subsection{Metric-Consistent Geometric Prior Integration}
\label{sec:ff_geometry}
Pretrained visual models provide camera and depth estimates from RGB images, but their geometries are inconsistent at three levels. In scale, VGGT-$\Omega$~\cite{ff_omega} predicts relative depth within each overlapping temporal window, whereas \mbox{MoGe-2}~\cite{ff_moge2} provides image-wise metric depth. In coordinate frame, each VGGT-$\Omega$ window has its own reference. In image grid, the predicted intrinsics refer to each model's input resolution. Left unresolved, the scale and frame inconsistencies misregister Gaussians from different context frames as well as target cameras; Sec.~\ref{sec:ff_ablation} quantifies the effect of scale. We resolve them in turn: metric calibration fixes scale, cross-window alignment fixes coordinate frames, and intrinsic conversion aligns camera rays with the reconstruction grid.

\textbf{Metric calibration.}
Since overlapping windows share images, the MoGe-2 depth of a shared anchor image provides a common metric reference for all windows that contain it. We therefore relate the relative depth of each window to the metric depth through a window-specific scale $s_w$ and an image-specific offset $b_i$ shared across windows:
\begin{equation}
    M_i(p)\approx s_wD_{wi}(p)+b_i,\qquad s_w>0,
\end{equation}
where $D_{wi}(p)$ is the relative depth at pixel $p$ of image $i$ in window $w$, and $M_i(p)$ is the corresponding metric depth. All scales and offsets are estimated jointly by minimizing

\begin{equation}
\min_{\{s_w,b_i\}}
\sum_w \frac{1}{|\mathcal A_w|}
\sum_{i\in\mathcal A_w} E_{wi}(s_w,b_i)
+\lambda_b \sum_i \left(\frac{b_i}{q_i}\right)^2,
\label{eq:ff_shared_affine}
\end{equation}
where $\mathcal A_w$ is the set of anchor images selected in window $w$, $E_{wi}$ is a Huber loss on normalized depth residuals averaged over sampled valid pixels, and the median metric depth $q_i$ normalizes the offset penalty weighted by $\lambda_b$. Averaging within each image and window compensates for unequal numbers of valid pixels, while the shared offsets couple windows that observe the same image. The window scales are bounded by empirical depth-ratio intervals for robust fitting. The offsets absorb per-image biases of monocular depth and thus prevent them from biasing the window scales; only the fitted scales are applied to camera translations, leaving the depth supervision unchanged.

\textbf{Cross-window alignment.}
The scaled windows are then registered into a common scene frame through rigid $\mathrm{SE}(3)$ transformations estimated from overlapping cameras, which preserve the metric scale. Camera estimates shared by multiple windows are fused, and the predicted intrinsics are adjusted for image resizing, cropping, and padding, aligning the camera rays with the reconstruction image grid, onto which RGB, depth, and track masks are also resampled. The resulting poses, rays, and metric depth targets serve as consistent conditions for Gaussian reconstruction and cross-view object association.

\subsection{Geometry-Guided Dynamic Reconstruction}
\label{sec:ff_hierarchical_dynamics}
\textbf{Feed-forward Gaussian prediction.}
Given the geometrically consistent camera estimates from Sec.~\ref{sec:ff_geometry}, we employ a Transformer-based reconstructor that encodes image patches, Pl\"ucker ray maps, and timestamps into tokens~\cite{ff_storm}. To aggregate information across context frames, the scene tokens are recurrently updated~\cite{ff_ufo} and decoded into pixel-aligned 3D Gaussians. Each Gaussian $j$ originates from a context frame at time $\tau_j$, with its initial center defined as $\boldsymbol\mu_j=\mathbf o_j+d_j\mathbf r_j$, where $\mathbf o_j$ and $\mathbf r_j$ are the camera ray origin and direction, and $d_j$ denotes the predicted camera-frame depth. The Gaussian head additionally predicts scale, orientation, opacity, color, and lifespan~\cite{ff_pvg}. We retain Gaussians from all context frames to preserve complementary surface observations across views and time, providing the basis for subsequent motion modeling and temporal fusion. These components follow prior designs~\cite{ff_storm,ff_ufo,ff_pvg}; Sec.~\ref{sec:ff_ablation} ablates our geometry conditions and the object-level dynamics below.

\textbf{Geometry-guided cross-view object association.}
Grounding DINO~\cite{ff_groundingdino} and SAM~2~\cite{ff_sam2} produce 2D object tracks independently for each camera. Since the decoded Gaussian centers lie in the common scene frame of Sec.~\ref{sec:ff_geometry}, we associate these tracks in 3D. Specifically, for each track at each context time, we compute its 3D center as the coordinate-wise median of the Gaussian centers within its mask. Tracks from different cameras are compared by the median distance between their 3D centers over shared context times, and mutual nearest neighbors within a distance threshold form a group, with at most one track per camera. Each Gaussian $j$ inherits the group ID $g_j$ of the track covering its pixel, where positive IDs denote tracked foreground objects, including stationary ones, and zero denotes the background.

\textbf{Hierarchical motion modeling.}
To predict object motion, backbone features pooled within the masks of each group across views and context times form a temporal descriptor, from which a velocity MLP predicts a shared velocity $\mathbf v_g$ that couples the Gaussians of group $g$ across cameras. Cross-view identity is thus established by the predicted geometry, while motion is estimated from features.

While the shared velocity captures the overall translation of an object, it cannot represent spatially varying motion, such as that of a turning vehicle. We therefore refine the foreground Gaussian attributes, including the center $\widetilde{\boldsymbol\mu}_j$, and predict a residual velocity $\delta\mathbf v_j$, acceleration $\mathbf a_j$, and jerk $\mathbf j_j$ with a local motion head. The head takes patch features, pixel RGB values, refined centers, shared velocities, and normalized within-patch coordinates $\mathbf u_j$ as input. Following the higher-order motion formulation of SLARM~\cite{ff_slarm}, the center of Gaussian $j$ at target time $t$ is given by
\begin{equation}
\boldsymbol\mu_j(t)
=\widetilde{\boldsymbol\mu}_j
+(\mathbf v_{g_j}+\delta\mathbf v_j)\Delta t_j
+\tfrac12\mathbf a_j\Delta t_j^2
+\tfrac16\mathbf j_j\Delta t_j^3,
\label{eq:ff_source_trajectory}
\end{equation}
where $\Delta t_j=t-\tau_j$ is the time offset from the source timestamp, measured in seconds. The shared velocity models object-level translation, while the local terms capture spatially varying and non-uniform motion. Only foreground Gaussian centers evolve according to Eq.~\eqref{eq:ff_source_trajectory}; their scales, orientations, and colors remain fixed, while background centers remain stationary. Opacities are modulated during rendering as described below.

\textbf{Source-aware temporal fusion.}
Gaussians reconstructed from different context frames capture complementary surfaces of dynamic objects. We propagate these Gaussians to the target time using Eq.~\eqref{eq:ff_source_trajectory} and adaptively weight their contributions according to the temporal distance from their source observations. To account for varying object visibility across context frames, the weights are normalized over the observation times of each object. Formally, let $\mathcal S_g$ denote the context times at which object $g$ is observed in at least one camera. The weight for source time $s\in\mathcal S_g$ at target time $t$ is defined as
\begin{equation}
 w_{g,s}(t)=
 \frac{\exp[-(t-s)^2/(2\sigma^2)]}
      {\sum_{u\in\mathcal S_g}\exp[-(t-u)^2/(2\sigma^2)]},
\label{eq:ff_source_weights}
\end{equation}
where $\sigma=0.35\,\mathrm{s}$. Each weight scales the opacity $\widetilde{\alpha}_j$ of every Gaussian $j$ of object $g$ from source time $s$, i.e., $\alpha_j(t)=\widetilde{\alpha}_j\,w_{g_j,\tau_j}(t)$, so that contributions are normalized across observation times. Background Gaussian opacities follow lifespan-based decay~\cite{ff_pvg}. The target image is rendered through differentiable Gaussian rasterization~\cite{ff_gaussians}, with uncovered pixels composited using a ray-conditioned sky color~\cite{ff_storm}.

\subsection{Training Objectives}
\label{sec:ff_training_objectives}
The reconstruction backbone and dynamic heads are jointly trained through target-view rendering. We employ a photometric MSE loss $\mathcal L_{\mathrm{rgb}}$, assigning five times higher weights to pixels within object track masks to emphasize object regions. A full-image perceptual loss $\mathcal L_{\mathrm{LPIPS}}$~\cite{ff_lpips} further improves rendering fidelity. For geometric supervision, $\mathcal L_d$ measures the discrepancy between rendered depth and MoGe-2 pseudo-depth, with both depth maps normalized by the maximum valid pseudo-depth. We additionally introduce a motion regularization loss $\mathcal L_m$ that penalizes excessive Gaussian-specific residual velocity, acceleration, and jerk, encouraging local motion to remain consistent with the shared object-level motion. The overall objective is
\begin{equation}
 \mathcal L=\mathcal L_{\mathrm{rgb}}
   +0.05\mathcal L_{\mathrm{LPIPS}}
   +0.1\mathcal L_d+10^{-4}\mathcal L_m.
\label{eq:ff_training_loss}
\end{equation}

%% file: sec/experiments.tex
\section{Experiments}
\label{sec:ff_experiments}
\providecommand{\ffcheck}{\ensuremath{\checkmark}}
\providecommand{\ffcross}{\ensuremath{\times}}

\subsection{Experimental Setup}
\textbf{Datasets and evaluation protocol.}
We conduct our primary experiments on the Waymo Open Dataset~\cite{ff_waymo} and further examine cross-dataset adaptation through RGB-only fine-tuning on nuScenes~\cite{ff_nuscenes} and Argoverse2~\cite{ff_argoverse2}. On Waymo, we evaluate all 202 validation scenes using the front-left, front, and front-right cameras at 10\,Hz. Following the temporal sampling protocol of STORM~\cite{ff_storm}, each 2s clip uses frames $\{0,5,10,15\}$ as context and evaluates 12 interpolated timestamps $\{1\!:\!4,6\!:\!9,11\!:\!14\}$ and four extrapolated timestamps $\{16\!:\!19\}$, yielding 48 target images per complete three-view clip. The evaluation comprises 96,360 target images at $160\times240$ resolution (supplementary Sec.~D). As in pose-free evaluation~\cite{ff_anysplat,ff_offthegrid}, target cameras are estimated from the images: ours come from the same full-video front end as the context cameras (Sec.~\ref{sec:ff_geometry}), so target frames inform these conditions but never enter the reconstructor (supplementary Sec.~E).

\textbf{Metrics and baselines.}
We report PSNR and SSIM on full images and on dynamic regions defined by Waymo 3D box annotations following STORM~\cite{ff_storm}, averaging image-wise scores over the evaluation set. Geometric accuracy is measured by depth RMSE (D-RMSE) at pixels with valid LiDAR measurements, without scale alignment. Dataset cameras, LiDAR depth, and 3D box annotations are used only for evaluation.

We compare with pose-conditioned feed-forward methods STORM~\cite{ff_storm}, SLARM-F~\cite{ff_slarm}, UFO~\cite{ff_ufo}, and PointForward~\cite{ff_pointforward}, as well as pose-free methods DGGT~\cite{ff_dggt} and StreetForward~\cite{ff_streetforward}. STORM and SLARM-F are locally reproduced under matched evaluation settings (supplementary Sec.~B); results for other methods are taken from their respective publications unless stated otherwise.

\textbf{Implementation details.}
We train the three-view 2s model for 100k iterations; the 8s model is initialized from the 2s checkpoint and trained for another 50k iterations on NVIDIA H200 GPUs (details in supplementary Sec.~A.5). Given precomputed RGB-derived conditions, the 2s model reconstructs a three-view clip in 0.21\,s and renders each target timestamp in 13\,ms on a single RTX 4090.

\subsection{Novel-Time Synthesis on Waymo}
\label{sec:ff_waymo_2s}

\begin{table*}[t]
\centering
\caption{\textbf{Waymo novel-time synthesis.} Each clip contains four context timestamps from three cameras and 16 target timestamps (12 interpolated and 4 extrapolated; per-type results in supplementary Sec.~D). \emph{Train/Test} indicate whether the method avoids dataset-provided cameras, LiDAR depth, and 3D box annotations in the respective phase.}
\label{tab:waymo_2s}
\setlength{\tabcolsep}{4.2pt}
\renewcommand{\arraystretch}{1.13}
\footnotesize
\begin{tabular*}{\textwidth}{@{\extracolsep{\fill}}lccccccc@{}}
\toprule
Methods & \multicolumn{2}{c}{Annotation-free} & \multicolumn{2}{c}{Dynamic regions} & \multicolumn{2}{c}{Full image} & Depth \\
\cmidrule(lr){2-3}\cmidrule(lr){4-5}\cmidrule(lr){6-7}\cmidrule(l){8-8}
 & Train & Test & PSNR$\uparrow$ & SSIM$\uparrow$ & PSNR$\uparrow$ & SSIM$\uparrow$ & D-RMSE (m)$\downarrow$ \\
\midrule
\multicolumn{8}{@{}l}{\emph{Pose-conditioned feed-forward methods}} \\
STORM$^\dagger$ & \ffcross & \ffcross & 22.04 & 0.623 & 25.85 & 0.803 & 5.47 \\
UFO & \ffcross & \ffcross & -- & -- & 27.26 & 0.825 & 5.45 \\
SLARM-F$^\dagger$ & \ffcross & \ffcross & 22.95 & 0.672 & 26.98 & 0.827 & 4.78 \\
PointForward & \ffcross & \ffcross & \textbf{25.01} & \underline{0.768} & \textbf{28.48} & \textbf{0.861} & 3.35 \\
\midrule
\multicolumn{8}{@{}l}{\emph{Pose-free feed-forward methods}} \\
DGGT & \ffcross & \ffcheck & 22.80 & 0.651 & \textit{27.41} & \underline{0.846} & 3.47 \\
StreetForward & \ffcross & \ffcheck & \textit{24.30} & \textbf{0.827} & 27.01 & 0.818 & 3.14 \\
\midrule
\textbf{Ours} & \ffcheck & \ffcheck & \underline{24.36} & \textit{0.729} & \underline{27.66} & \textit{0.836} & 7.06 \\
\bottomrule
\end{tabular*}
\vspace{3pt}
\parbox{\textwidth}{\scriptsize
\ffcheck: without dataset-provided cameras, LiDAR depth, or 3D box annotations;
\ffcross: with any of them. $^\dagger$ denotes reproduced results; other results are reported
in their respective publications under their own protocols, except DGGT's dynamic-region results, which are from PointForward~\cite{ff_pointforward}; supplementary Sec.~E evaluates ours under the protocol of the DGGT code. Bold/underline/italic: first/second/third best in RGB metrics. --: unavailable.}
\end{table*}

\textbf{Quantitative comparison.}
Pose-conditioned STORM, SLARM-F, and UFO require dataset cameras as input; without them, FreeForward4D achieves 1.81, 0.68, and 0.40\,dB higher full-image PSNR, using frozen priors reconciled into a single metric frame instead. On dynamic regions, the gains over STORM and SLARM-F increase to 2.32 and 1.41\,dB in PSNR, with higher SSIM (0.729 vs.\ 0.623 and 0.672), consistent with our object-level motion modeling.

Among pose-free methods, FreeForward4D exceeds DGGT and StreetForward by 0.25 and 0.65\,dB in full-image PSNR, although both are trained with dataset annotations, and remains higher in PSNR under the evaluation protocol of the official DGGT code (supplementary Sec.~E). Only PointForward, which uses ground-truth 3D box trajectories, attains higher PSNR on both full images and dynamic regions.

\begin{figure*}[t]
\centering
\includegraphics[width=\textwidth]{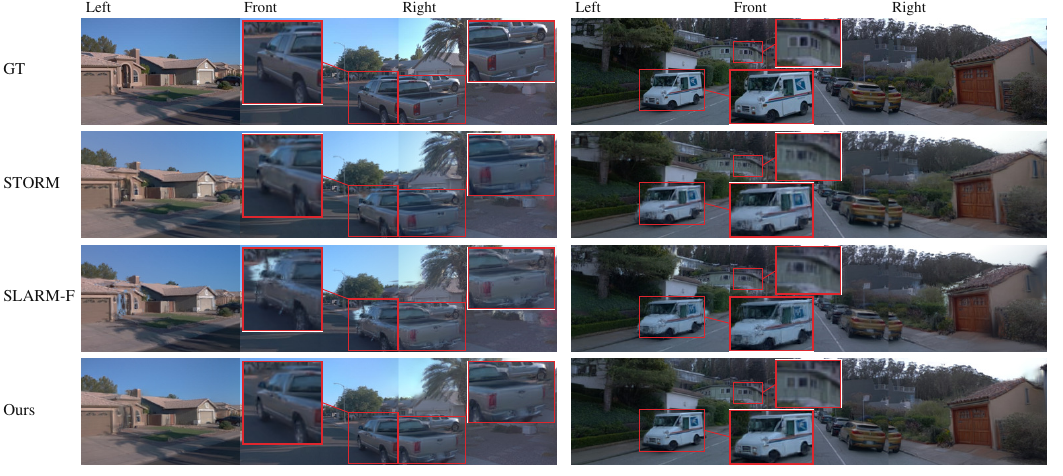}
\caption{\textbf{Waymo qualitative comparison at extrapolated timestamps.} Each example shows the left, front, and right cameras; rows correspond to ground truth, STORM, SLARM-F, and ours. Enlarged regions illustrate vehicle boundaries, temporal ghosting, and fine scene details. The baselines require dataset cameras; ours renders at cameras recovered by its own front end.}
\label{fig:ff_qualitative}
\end{figure*}

\textbf{Qualitative comparison.}
Figure~\ref{fig:ff_qualitative} compares extrapolated timestamps. FreeForward4D reconstructs sharper vehicle boundaries and finer appearance details than STORM and SLARM-F while preserving static structures, without dataset cameras.

\subsection{Metric Geometry and Motion from RGB}
\label{sec:ff_scene_level}

\begin{figure}[t]
\centering
\includegraphics[trim=2.64 0 340.76 0,clip,width=0.495\columnwidth]{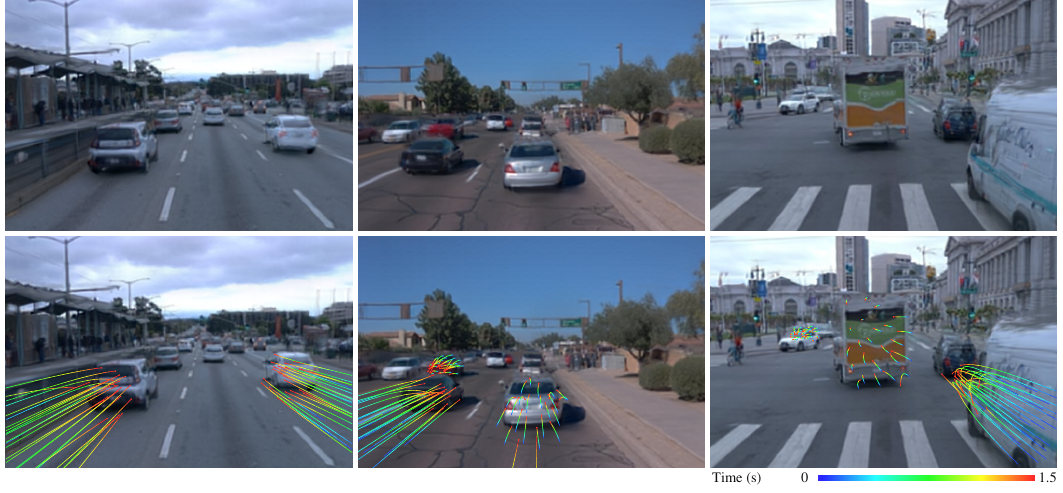}\hfill
\includegraphics[trim=340.8 0 2.6 0,clip,width=0.495\columnwidth]{figures/tracking_bev_1.pdf}
\caption{\textbf{Novel-view rendering and Gaussian trajectories on Waymo.} Top: renderings from virtual cameras elevated by 0.5\,m relative to the input views. Bottom: time-colored trajectories of selected Gaussians on moving vehicles over 0--1.5\,s.}
\label{fig:ff_tracking}
\end{figure}

\paragraph{Metric geometry evaluation.}
FreeForward4D is the only method in Table~\ref{tab:waymo_2s} that uses no dataset cameras, LiDAR, or alignment in either training or evaluation, drawing metric scale from the frozen MoGe-2 prior, whose training data include Waymo LiDAR (supplementary Sec.~A): pose-conditioned methods inherit it from dataset cameras and mostly LiDAR, DGGT aligns its depth to LiDAR before evaluation~\cite{ff_dggt}, and StreetForward is optionally supervised with ground-truth depth. Its D-RMSE of 7.06\,m on Waymo 2s therefore jointly reflects scale, camera, and depth errors.

Camera accuracy tests the recovered scale directly: each 2s clip is anchored at its first front-camera pose without scale adjustment or trajectory fitting. Across 1,818 clips, the absolute trajectory error is 0.58\,m in median (1.08\,m mean, 2.55\,m P90) over a median ground-truth path length of 11.0\,m, while the median relative translation error over 0.5\,s is 0.27\,m and the median rotation error is $1.20^\circ$ (supplementary Sec.~C). Figure~\ref{fig:ff_tracking} visualizes both aspects: renderings from cameras elevated by 0.5\,m preserve road layout and vehicle shapes, and Gaussians on the same vehicle follow parallel, smooth trajectories.

\subsection{Extended 8s Reconstruction}
We evaluate extended temporal windows on 8-second sequences using 16 context frames sampled at 0.5-second intervals. As shown in Table~\ref{tab:ff_8s}, our method achieves the best RGB rendering performance on Waymo, with 27.57\,dB PSNR and 0.836 SSIM, surpassing UFO by 0.18\,dB and STORM by 3.09\,dB. Evaluated without alignment, our D-RMSE (7.11\,m) is lower than those of STORM (8.11\,m) and GS-LRM (7.37\,m) and above that of UFO (5.10\,m), all of which use dataset annotations. It also merges static Gaussians over about 20\,s without per-scene optimization (supplementary Sec.~F).

\begin{table}[t]
\centering
\footnotesize
\setlength{\tabcolsep}{0pt}
\renewcommand{\arraystretch}{1.06}
\caption{\textbf{Extended 8s reconstruction.}
Each Waymo window uses 16 context frames
and 64 target timestamps. Baseline results are reported
by UFO~\cite{ff_ufo}. Results on nuScenes and Argoverse2
are obtained through RGB-only fine-tuning of our
Waymo-trained 8s model.}
\label{tab:ff_8s}
\begin{tabular*}{\columnwidth}{@{\extracolsep{\fill}}lccccc@{}}
\toprule
 & \multicolumn{3}{c}{\textbf{Rendering and depth quality}}
 & \textbf{Feed-} & \textbf{Annot.-} \\
\cmidrule(lr){2-4}
\textbf{Method}
& \textbf{PSNR}$\uparrow$
& \textbf{SSIM}$\uparrow$
& \textbf{D-RMSE (m)}$\downarrow$
& \textbf{forward}
& \textbf{free} \\
\midrule
\multicolumn{6}{@{}l}{\textit{Waymo Open}} \\
3DGS~\cite{ff_gaussians} & 19.57 & 0.517 & 14.42 & \ffcross & \ffcross \\
PVG~\cite{ff_pvg} & 22.90 & 0.619 & 18.24 & \ffcross & \ffcross \\
Street Gaussians~\cite{ff_streetgaussians} & 21.69 & 0.609 & 13.17 & \ffcross & \ffcross \\
DeformableGS~\cite{ff_deformablegs} & 21.47 & 0.611 & 13.12 & \ffcross & \ffcross \\
GS-LRM~\cite{ff_gslrm} & 21.81 & 0.584 & 7.37  & \ffcheck & \ffcross \\
STORM~\cite{ff_storm} & 24.48 & 0.736 & 8.11  & \ffcheck & \ffcross \\
UFO~\cite{ff_ufo} & \underline{27.39} & \underline{0.830}
                  & 5.10 & \ffcheck & \ffcross \\
\midrule
\textbf{Ours}     & \textbf{27.57} & \textbf{0.836}
                  & 7.11 & \ffcheck & \ffcheck \\
\midrule
\multicolumn{6}{@{}l}{\textit{Cross-dataset adaptation}} \\
Ours (nuScenes)   & 25.70 & 0.803 & -- & \ffcheck & \ffcheck \\
Ours (Argoverse2)& 24.92 & 0.811 & -- & \ffcheck & \ffcheck \\
\bottomrule
\end{tabular*}
\end{table}

\subsection{Cross-Dataset Adaptation}
\label{sec:ff_adaptation}
Because FreeForward4D requires no annotations, adapting it to a new dataset or camera rig requires only raw videos. Whereas the baselines are trained on each target dataset with its annotations, we only fine-tune the Waymo-trained 2s model on nuScenes and Argoverse2 using RGB-derived conditions, with 700 training and 150 validation scenes per dataset and the temporal sampling used on Waymo.

As shown in Table~\ref{tab:ff_cross_dataset}, RGB-only fine-tuning alone reaches 26.90\,dB PSNR on nuScenes, 0.27\,dB above DGGT and 2.36\,dB above STORM, both trained on nuScenes with dataset annotations. On Argoverse2, it attains 26.52\,dB, 1.55\,dB above STORM and within 0.44\,dB of the annotation-trained DGGT. Fine-tuning the 8s model in the same way reaches 25.70\,dB on nuScenes and 24.92\,dB on Argoverse2 (Table~\ref{tab:ff_8s}).

\begin{table}[t]
\centering
\footnotesize
\setlength{\tabcolsep}{3pt}
\renewcommand{\arraystretch}{1.12}
\caption{\textbf{Cross-dataset adaptation on nuScenes and Argoverse2.}
Full-image novel-time synthesis after adaptation to each target dataset.
\emph{Train/Test} indicate whether the method avoids dataset-provided cameras, LiDAR depth, and 3D box annotations in the respective phase.
Ours is fine-tuned from the Waymo model using only RGB-derived conditions, whereas the baselines are trained on each target dataset with dataset annotations; baseline results are from their publications, with STORM as reported by DGGT~\cite{ff_dggt}.}
\label{tab:ff_cross_dataset}
\begin{tabular*}{\columnwidth}{@{\extracolsep{\fill}}lcccccc@{}}
\toprule
Methods & \multicolumn{2}{c}{Annotation-free}
& \multicolumn{2}{c}{nuScenes}
& \multicolumn{2}{c}{Argoverse2} \\
\cmidrule(lr){2-3}\cmidrule(lr){4-5}\cmidrule(l){6-7}
& Train & Test
& PSNR$\uparrow$ & SSIM$\uparrow$
& PSNR$\uparrow$ & SSIM$\uparrow$ \\
\midrule
STORM~\cite{ff_storm}
& \ffcross & \ffcross
& 24.54 & 0.784
& 24.97 & 0.791 \\
DGGT~\cite{ff_dggt}
& \ffcross & \ffcheck
& 26.63 & 0.813
& 26.96 & 0.831 \\
\midrule
\textbf{Ours}
& \ffcheck & \ffcheck
& 26.90 & 0.813
& 26.52 & 0.829 \\
\bottomrule
\end{tabular*}

\end{table}

\begin{table}[t]
\centering\footnotesize
\setlength{\tabcolsep}{0pt}\renewcommand{\arraystretch}{1.06}
\caption{\textbf{Ablation study on Waymo 2s.} \emph{w/o geom. integ.} uses, for each clip, the uncalibrated VGGT-$\Omega$ cameras of the single window covering it; \emph{w/o object motion} removes grouping, foreground refinement, and shared velocity; \emph{w/o local motion} retains shared object velocity only; \emph{w/o accel.\ \& jerk} retains first-order residual motion; \emph{w/o normalization} removes per-object normalization of temporal weights.}
\label{tab:ff_ablation}
\begin{tabular*}{\columnwidth}{@{\extracolsep{\fill}}lccccc@{}}
\toprule
 & \multicolumn{2}{c}{\textbf{Full image}} & \multicolumn{2}{c}{\textbf{Dynamic regions}} & \textbf{Depth} \\
\cmidrule(lr){2-3}\cmidrule(lr){4-5}\cmidrule(l){6-6}
\textbf{Variant} & \textbf{PSNR}$\uparrow$ & \textbf{SSIM}$\uparrow$ & \textbf{PSNR}$\uparrow$ & \textbf{SSIM}$\uparrow$ & \textbf{D-RMSE (m)}$\downarrow$ \\
\midrule
w/o geom. integ. & 23.73 & 0.677 & 22.57 & 0.608 & 7.99 \\
\addlinespace[2pt]
w/o object motion & 27.26 & 0.829 & 22.77 & 0.651 & 7.22 \\
w/o cross-view assoc. & 27.62 & \underline{0.835} & 24.26 & 0.726 & 7.06 \\
w/o local motion & 27.55 & 0.833 & 24.12 & 0.719 & 7.06 \\
w/o accel. \& jerk & \underline{27.64} & \underline{0.835} & \underline{24.30} & \underline{0.727} & \underline{7.05} \\
\addlinespace[2pt]
w/o normalization & 27.59 & \underline{0.835} & 24.04 & 0.724 & \textbf{7.01} \\
\midrule
Ours & \textbf{27.66} & \textbf{0.836} & \textbf{24.36} & \textbf{0.729} & 7.06 \\
\bottomrule
\end{tabular*}
\end{table}

\subsection{Ablation Studies}
\label{sec:ff_ablation}
We ablate each component on Waymo 2s (Table~\ref{tab:ff_ablation}) by disabling it in the trained full model at inference, so all variants share the same weights and differ only in the disabled component.

\textbf{Geometric prior integration.}
Uncalibrated single-window cameras cause the largest drop in Table~\ref{tab:ff_ablation}: 3.93\,dB in full-image PSNR and 1.79\,dB on dynamic regions. As a 2s clip lies within one window, this variant keeps a consistent coordinate frame but lacks the metric scale: camera translations in relative units no longer match the metric depths, misregistering Gaussians from different context frames and target cameras (Fig.~\ref{fig:ff_pipeline}b). D-RMSE rises from 7.06 to 7.99\,m. Cross-window alignment instead registers longer sequences, as in the 8s setting and scene-level merging (supplementary Sec.~F).

\textbf{Hierarchical motion modeling.}
Removing object motion reduces dynamic-region PSNR by 1.59\,dB but full-image PSNR by only 0.40\,dB. Disabling only cross-camera association lowers dynamic-region PSNR from 24.36 to 24.26\,dB. Of these 1.59\,dB, object-level motion (cross-view grouping with a shared velocity) contributes 1.35\,dB (22.77 to 24.12\,dB), as a single object motion explains most dynamics within 2\,s; the local motion head adds the remaining 0.24\,dB through foreground refinement with residual velocity (24.30\,dB) and acceleration and jerk (24.36\,dB) for spatially varying and non-uniform motion. Without any 3D annotation, grouping reaches a pairwise precision/recall of 0.867/0.647 on label-matched tracks (supplementary Sec.~A.3).

\textbf{Source-aware temporal fusion.}
Without per-object normalization, dynamic-region PSNR drops from 24.36 to 24.04\,dB while D-RMSE barely changes. Without it, an object far from all its observations, e.g., at extrapolated times, fades; normalization preserves its opacity while favoring the nearest observation.

%% file: sec/Conclusion.tex
\section{Conclusion}
\label{sec:conclusion}
We presented FreeForward4D, an annotation-free feed-forward framework that reconstructs dynamic driving scenes from RGB videos. Reconciling frozen visual priors into a single metric frame enables cross-view object association and shared object motion without 3D box annotations; replacing the calibrated cameras with uncalibrated ones costs 3.93\,dB in full-image PSNR. Without dataset annotations, FreeForward4D surpasses STORM, SLARM-F, UFO, DGGT, and StreetForward in full-image PSNR on Waymo and adapts to nuScenes and Argoverse2 through RGB-only fine-tuning. Its depth accuracy and dynamics are bounded by the frozen priors; needing no annotations, it can directly benefit from stronger priors and more unannotated videos.

%% file: sec/supplement.tex
\clearpage
\setcounter{page}{1}
\maketitlesupplementary
\appendix
\renewcommand{\theequation}{S\arabic{equation}}
\renewcommand{\thetable}{S\arabic{table}}
\renewcommand{\thefigure}{S\arabic{figure}}
\section{Implementation Details}
This section details the calibration, rendering, and training used by FreeForward4D.

\paragraph{Pretrained priors.}
FreeForward4D uses four publicly released models, all kept frozen and never fine-tuned on our data: VGGT-$\Omega$~\cite{ff_omega} for relative cameras and depth, MoGe-2~\cite{ff_moge2} for monocular metric depth, Grounding DINO~\cite{ff_groundingdino} for open-vocabulary detection, and SAM~2~\cite{ff_sam2} for mask propagation. Their training data are described in the respective papers; in particular, MoGe-2 is trained on a mixture that includes LiDAR depth from driving datasets such as Waymo and Argoverse2~\cite{ff_moge2}, without a specified split. FreeForward4D itself uses no camera, LiDAR, or box annotation of any dataset for training or inference.

\subsection{Shared Affine Metric Calibration}
\paragraph{Anchors and valid support.}
On all datasets, VGGT-$\Omega$ processes overlapping windows of 20 consecutive timestamps with three frontal camera views per timestamp (60 images per window); the temporal stride is one timestamp on Waymo and five on nuScenes and Argoverse2. Following~\cite{chen2025spatial}, we calibrate each window against monocular metric depth; overlapping windows are additionally coupled by shared per-image offsets. Each window selects up to three confidence-ranked anchor images. A fitting pixel must have finite relative depth $D_{wi}(p)>10^{-4}$, finite metric depth $M_i(p)>10^{-4}$, a valid MoGe-2 mask, and VGGT-$\Omega$ confidence above the image median. An anchor observation with fewer than 100 such pixels is discarded. From each surviving observation we sample at most 2,048 evenly spaced entries of its valid-pixel list, forming $\mathcal P_{wi}$; $\mathcal A_w$ is the surviving anchor set. The reference $q_i$ is the median metric depth over valid MoGe-2 pixels before window-specific confidence filtering.

\paragraph{Fitting error.}
For positive window scale $s_w$ and per-image offset $b_i$, the main paper's fitting term is
\begin{equation}
\begin{aligned}
 E_{wi}(s_w,b_i)
 &=\frac{1}{|\mathcal P_{wi}|}\sum_{p\in\mathcal P_{wi}}
   \rho\!\left(\frac{s_wD_{wi}(p)+b_i-M_i(p)}{\epsilon_i(p)}\right),\\
 \epsilon_i(p)&=0.05\max\!\left(M_i(p),0.05q_i\right).
\end{aligned}
\label{eq:supp_affine_error}
\end{equation}
Here $\rho$ is the Huber loss with threshold 1.5: $\rho(r)=r^2/2$ for $|r|\le1.5$ and $1.5(|r|-0.75)$ otherwise. The offset coefficient is $\lambda_b=200$. Thus the penalty is $200\sum_i(b_i/q_i)^2$, and the data term averages pixels within anchors and anchors within windows. All quantities are computed from frozen RGB predictions.

\paragraph{Scale bounds and optimization.}
Let $\mathcal R_w$ pool valid ratios $M_i(p)/D_{wi}(p)$ from the surviving anchors \emph{before} pixel subsampling. Each window scale is constrained to the 40th--60th percentile interval of $\mathcal R_w$, which guards against degenerate fits on windows with weak depth support. The initial scale is the median of sampled ratios, projected onto this interval. Iteratively reweighted least squares uses Schur-complement elimination of the shared offsets and projected scale updates. Only the fitted scales modify camera translations, expressed relative to each window's reference camera. The offsets are used only during calibration, and metric depth targets remain unshifted.

\subsection{Camera Alignment and Image Grids}
An incoming window requires at least nine images shared with the already aligned windows. The alignment rotation is the $\mathrm{SO}(3)$ projection of the sum of relative camera rotations. Translation is the coordinate-wise median of the destination camera centers minus the rotated source centers. Pairs pass separate rotation and translation tests at the median residual plus three median absolute deviations; a $10^{-6}$ numerical offset is added to each deviation scale. We refit when at least nine pairs remain. These rigid $\mathrm{SE}(3)$ transformations preserve the estimated metric scale. Repeated poses use rotation averaging and median translation, and repeated intrinsics use elementwise medians.

Let $A_{wi}^{\Omega}$ map original image coordinates to the prior grid, and $A_i^{\mathrm{rec}}$ map them to the reconstruction grid. The intrinsic matrix $K_{wi}^{\Omega}$ predicted on the prior grid is converted by
\begin{equation}
 \widehat K_{wi}=A_i^{\mathrm{rec}}(A_{wi}^{\Omega})^{-1}K_{wi}^{\Omega}.
\label{eq:supp_intrinsics}
\end{equation}
The transforms include resizing, cropping, and padding. RGB, metric depth, and object masks use the reconstruction pixel grid; track identities are resampled with nearest-neighbor interpolation.

\subsection{Gaussian Construction and Object Motion}
Each $8\times8$ image patch produces 64 pixel-aligned Gaussians through a two-layer decoder. The ray used for the initial center has camera-frame $z=1$ and is transformed to scene coordinates without unit-length normalization. Consequently the predicted depth is camera-frame $z$-depth. Unit directions are used for the Pl\"ucker encoding. Each Gaussian retains its context time.

\paragraph{2D object tracks.}
Tracks are computed offline, independently for each camera over the full video with Grounding DINO~\cite{ff_groundingdino} and SAM~2~\cite{ff_sam2}. Every 10 frames, Grounding DINO detects objects with the prompt ``car. truck. bus. motorcycle. bicycle. pedestrian.'' (box and text thresholds 0.25), and SAM~2 converts the detected boxes into masks. A new mask inherits the ID of an existing track when their mask IoU is at least 0.8 and otherwise starts a new track. Masks are propagated forward for 10 frames, and objects first detected at a keyframe are additionally propagated backward for up to 20 frames. Consequently, objects first detected at a later keyframe obtain masks at earlier context frames from frames up to 2\,s later, as permitted by the offline setting of the main paper (Sec.~3). Only the masks and track identities at the selected context timestamps are passed to the reconstructor; target-frame masks are used only to weight the training loss.

\paragraph{Cross-view association.}
A track observation contributes a 3D center only when its mask contains at least 16 finite Gaussian centers. Coordinate-wise medians are detached before association. Cross-camera mutual nearest neighbors are accepted when their median Euclidean distance over common context times is at most $4\,\mathrm{m}$ and they share at least one context time. A group contains at most one track per camera. Tracks without a valid center keep distinct positive IDs.

We evaluate association against Waymo 3D labels, used for evaluation only. Each track is assigned to a labeled object by 2D box overlap (IoU $\geq 0.3$) on context frames, which matches 44.6\% of the 83,701 tracks over 2,020 windows. Among cross-camera pairs of matched tracks that share a context time, the $4\,\mathrm{m}$ threshold gives a pairwise precision of 0.867 and recall of 0.647; varying it from 3 to $6\,\mathrm{m}$ changes the F1 score only from 0.720 to 0.751. Centers of the same object seen by different cameras differ by a median of $1.53\,\mathrm{m}$, and 83.2\% of such pairs lie within $4\,\mathrm{m}$, compared with 1.4\% of pairs from different objects. These values exclude unmatched tracks and include errors of the label assignment.

\paragraph{Shared and local heads.}
Mask coverage weights feature pooling over views at each context time. The shared-velocity MLP receives the temporal mean and least-squares temporal slope of pooled features, the observation duration, and $\log(1+\bar c)/5$, where $\bar c$ is the mean total patch coverage. Its output is bounded componentwise to $(-20,20)\,\mathrm{m/s}$. Groups observed at fewer than two context times receive zero shared velocity.

The attribute head uses patch features, input RGB, decoded color, and opacity to refine foreground centers, scales, quaternions, opacities, and colors once per reconstruction. The local head takes patch features, RGB, the refined center, shared velocity, and the normalized pixel position within the patch. Residual velocity, acceleration, and jerk have componentwise bounds $4\,\mathrm{m/s}$, $6\,\mathrm{m/s^2}$, and $12\,\mathrm{m/s^3}$, respectively. The temporal trajectory changes Gaussian centers; refined scale, orientation, and color remain fixed during target-time evaluation. Background centers remain fixed.

\subsection{Temporal Rendering}
All views of one frame share the assigned frame timestamp. For object $g$, $\mathcal S_g$ contains context times with at least one object pixel in any view. The normalized kernel of the main paper uses $\sigma=0.35\,\mathrm{s}$ and is broadcast to all Gaussians of the same group and context time. If $\tau_j$ is Gaussian $j$'s context time, rendering uses
\begin{equation}
\begin{aligned}
 \alpha_j^{\mathrm{fg}}(t)&=\widetilde\alpha_jw_{g_j,\tau_j}(t),\\
 \alpha_j^{\mathrm{bg}}(t)&=\alpha_j\exp\!\left[-\frac{(t-\tau_j)^2}{2\ell_j^2}\right].
\end{aligned}
\label{eq:supp_opacity}
\end{equation}
Here $\widetilde\alpha_j$ is refined foreground opacity, $\alpha_j$ is decoded background opacity, and $\ell_j>0$ is its predicted lifespan. The lifespan is clamped below at $10^{-6}$ for numerical evaluation. Foreground temporal weights sum to one over the observed context times; all views and Gaussians at one context time receive the same weight.

\subsection{Training Losses}
For target pixel $p$, let $\widehat{\mathbf I}_p$ and $\mathbf I_p$ be rendered and observed RGB. The binary mask $m_p$ indicates a positive SAM track ID. With $\omega_p=1+4m_p$,
\begin{equation}
 \mathcal L_{\mathrm{rgb}}=
 \frac{\sum_p\omega_p\|\widehat{\mathbf I}_p-\mathbf I_p\|_2^2}
      {3\sum_p\omega_p}.
\label{eq:supp_rgb}
\end{equation}
Full-image LPIPS supplies the perceptual loss. Let $\mathcal V$ contain valid MoGe-2 pixels with finite detached MoGe-2 depth $D_p^{\mathrm M}>0.01\,\mathrm{m}$. Pseudo-depth supervision is
\begin{equation}
\begin{aligned}
 \mathcal L_d&=\frac{1}{|\mathcal V|}\sum_{p\in\mathcal V}
       \frac{|\widehat D_p-D_p^{\mathrm M}|}{D_{\max}^{\mathrm M}},\\
 D_{\max}^{\mathrm M}&=\max_{p\in\mathcal V}D_p^{\mathrm M},
\end{aligned}
\label{eq:supp_depth}
\end{equation}
where $\widehat D_p$ is rendered camera-frame $z$-depth and the maximum is taken over the target views supervised together in one chunk. Foreground Gaussians form $\mathcal F$. The local coefficient penalty is
\begin{equation}
\begin{aligned}
 \mathcal L_m&=\frac{1}{3\max(|\mathcal F|,1)}
 \sum_{j\in\mathcal F}\bigl(\|\delta\mathbf v_j\|_2^2\\
 &\qquad+0.1\|\mathbf a_j\|_2^2+0.01\|\mathbf j_j\|_2^2\bigr).
\end{aligned}
\label{eq:supp_motion}
\end{equation}
If $\mathcal L_c$ is the main paper's four-term objective for target chunk $c$, the aggregate objective is
\begin{equation}
 \mathcal L_{\mathrm{train}}=\frac{4}{C}\sum_{c=1}^{C}\mathcal L_c.
\label{eq:supp_chunks}
\end{equation}
This keeps the loss scale of a four-chunk clip across clip lengths. The reconstruction backbone and dynamic heads are trained jointly; visual-prior and perceptual models remain frozen.

\paragraph{Optimization.}
The three-view 2s model is trained for 100k iterations with a global batch size of 64 and an initial learning rate of $10^{-4}$, using 5k warmup iterations and cosine decay to $10^{-6}$. The 8s model is initialized from the 2s checkpoint and trained for another 50k iterations with a batch size of 32 and a learning rate of $3\times10^{-5}$.

\section{Reproduced Baselines}
STORM$^\dagger$ and SLARM-F$^\dagger$ in Table~1 of the main paper are retrained with the authors' official open-source implementations without code modification and evaluated on the same 96,360 target images under the same protocol as ours. With our deduplicated 96,360 targets, STORM$^\dagger$ reaches 25.85\,dB full-image PSNR, compared with 26.38\,dB reported in~\cite{ff_storm}; FreeForward4D (27.66\,dB) exceeds both.

\section{Accuracy of the Estimated Cameras}
We score the estimated cameras on all 202 Waymo validation scenes against the dataset poses, which are used only here. Each 2s clip (the nine non-overlapping 20-frame windows of each scene, starting every 20 frames) is re-based at its first front-camera pose for both trajectories; no scale, rotation, or translation is fitted. ATE is the translation RMSE over the three cameras and 20 frames. RPE is the translation RMSE of 0.5\,s relative motions of the same camera. Rotation error is the geodesic angle between the re-based estimated and dataset camera rotations, averaged over the three cameras and 20 frames. Table~\ref{tab:supp_cameras} summarizes the results.

\begin{table}[h]
\centering\small
\setlength{\tabcolsep}{5pt}
\caption{\textbf{Camera accuracy on Waymo validation.} 1,818 2s clips across 202 scenes.}
\label{tab:supp_cameras}
\begin{tabular}{lrrr}
\toprule
Metric & Mean & Median & P90\\
\midrule
ATE, 2s clip (m) & 1.08 & 0.58 & 2.55\\
RPE, 0.5\,s (m) & 0.53 & 0.27 & 1.20\\
Rotation error ($^\circ$) & 1.52 & 1.20 & 2.47\\
\bottomrule
\end{tabular}
\end{table}
The median ground-truth path of a clip is 11.0\,m.

\section{Evaluation Protocols}
\paragraph{Waymo 2s.}
The 2s evaluation uses ten 20-frame windows per scene (2,020 in total): nine start every 20 frames, and the last is aligned to the end of the scene. Target frames shared with the preceding window are counted once, which yields 72,132 interpolated and 24,228 extrapolated images (96,360 in total). Table~\ref{tab:supp_interp_extrap} splits the Waymo 2s results of the main paper by target type. Interpolated targets are frames $\{1\!:\!4,6\!:\!9,11\!:\!14\}$ between context frames, and extrapolated targets are frames $\{16\!:\!19\}$ after the last context frame. All rows use the same model, cameras, and evaluation set as the main paper. Dynamic-region scores average over images containing dynamic pixels.

\begin{table}[h]
\centering\footnotesize
\setlength{\tabcolsep}{1.5pt}
\caption{\textbf{Waymo 2s results by target type.} Full-image and dynamic-region RGB fidelity and depth error.}
\label{tab:supp_interp_extrap}
\begin{tabular}{lrccccc}
\toprule
 & & \multicolumn{2}{c}{Full image} & \multicolumn{2}{c}{Dynamic regions} & Depth \\
\cmidrule(lr){3-4}\cmidrule(lr){5-6}\cmidrule(l){7-7}
Targets & Images & PSNR$\uparrow$ & SSIM$\uparrow$ & PSNR$\uparrow$ & SSIM$\uparrow$ & D-RMSE$\downarrow$ \\
\midrule
Interp. & 72,132 & 28.09 & 0.848 & 24.88 & 0.755 & 6.96 \\
Extrap. & 24,228 & 26.39 & 0.798 & 22.96 & 0.660 & 7.35 \\
\midrule
All & 96,360 & 27.66 & 0.836 & 24.36 & 0.729 & 7.06 \\
\bottomrule
\end{tabular}
\end{table}

\paragraph{Waymo 8s.}
Each 8s clip spans 80 frames at 10\,Hz from the same three cameras at $160\times240$ resolution. Frames $0,5,\dots,75$ serve as the 16 context frames, and the remaining 64 timestamps are targets: 60 interpolated and frames $\{76\!:\!79\}$ extrapolated. Following the protocol of UFO~\cite{ff_ufo}, clips start at frames $0,20,\dots,180$, and a start $s$ with $s+80\geq N$, where $N$ is the scene length, is replaced by a random start in $[0,N-80]$. This yields 2,019 clips over the 202 validation scenes and 387,648 target images; overlapping clips are not deduplicated.

\paragraph{nuScenes.}
We use the front-left, front, and front-right cameras. The 12\,Hz sweeps are resampled to a 10\,Hz grid anchored at the front camera with offsets of at most 60\,ms; each side camera takes its nearest exposure, and frames whose three exposures span more than 25\,ms are discarded. Images are center-cropped from $1600\times900$ to $1350\times900$ and resized to $160\times240$. Fine-tuning uses the 700 training scenes and evaluation the 150 official validation scenes. The 2s protocol follows Waymo, with context frames $\{0,5,10,15\}$ and clips starting every 20 frames, giving 1,350 clips and 64,800 target images. The 8s protocol uses clips starting at frames $0,20,\dots,100$, six per scene, giving 900 clips and 172,800 target images.

\paragraph{Argoverse2.}
We use the front-left, front-center, and front-right ring cameras. The 20\,Hz streams are resampled to a 10\,Hz grid anchored at the center camera, with offsets of at most 5\,ms for the center and 25\,ms for the side cameras; images are center-cropped to a 3:2 aspect ratio and resized to $160\times240$. Each of the 150 validation logs forms one sequence of 139--198 frames. The 2s protocol uses non-overlapping 20-frame windows with context frames $\{0,5,10,15\}$, giving 1,104 windows and 52,992 target images. The 8s protocol uses 80-frame windows with stride 20, giving 654 windows and 125,568 target images.

\section{Target Cameras}
\label{sec:supp_target_cameras}
\paragraph{Main-paper protocol.}
Following the pose-free evaluation of AnySplat~\cite{ff_anysplat} and Off The Grid~\cite{ff_offthegrid}, target cameras are estimated from the images while the scene is reconstructed from the context frames. Our front end processes the full video in a single pass, so the context cameras, the metric calibration, and the 2D object tracks are obtained with all frames available; target RGB never enters the reconstruction network, consistent with the offline setting of the main paper (Sec.~3). Rendering a pose-free reconstruction at dataset cameras would first require registering the dataset frame to the frame of the reconstruction, and errors of this registration would enter the image metrics; estimating target cameras in the frame of the reconstruction keeps the evaluation on rendering quality. Unlike AnySplat, which passes the target images through its network to predict their poses, our target frames inform only the front end. Comparing such renderings with pose-conditioned methods rendered at dataset cameras follows DGGT~\cite{ff_dggt}.

\paragraph{Protocol of the DGGT code.}
We also evaluate with the protocol of the official DGGT code~\cite{ff_dggt}. From the context frames $\{0,5,10,15\}$ of each clip, frames $0$--$15$ are rendered and the metrics are averaged over all 16, so the four context frames, which are inputs to the reconstruction, are scored together with the 12 intermediate frames, and the extrapolated frames $\{16\!:\!19\}$ are not evaluated. Target cameras are interpolated between the context cameras, with spherical interpolation for rotation and linear interpolation for translation, rather than estimated from the target images. We keep the clips and the $160\times240$ resolution of Table~1 of the main paper. Because the evaluated frames differ, these results are not directly comparable with Table~1 of the main paper. Table~\ref{tab:supp_dggt_protocol} reports the 16 frames together and the 12 intermediate frames alone. DGGT describes its evaluation as covering all frames, with inputs 0, 5, 10, and 15 and predictions for frames 0--19, and describes its reported novel-view synthesis results as computed on interpolated frames; its public code scores frames 0--15. On both frame sets, FreeForward4D attains higher PSNR than the 27.41\,dB reported by DGGT. On the same 12 intermediate frames, front-end target cameras give 28.09\,dB (Table~\ref{tab:supp_interp_extrap}) and interpolated cameras 27.57\,dB; the two differ only in the target cameras. Pose-conditioned baselines render at the exact dataset cameras of the target frames, which are consistent with their dataset context cameras; the counterpart for a pose-free method is to estimate the target cameras in its own frame. These estimates are not more accurate than the dataset cameras: their median trajectory error is 0.58\,m over 2s clips (Sec.~C). Our context cameras are estimated in the same way, so pose errors among them, which the baselines do not incur, affect only our reconstructions.

\begin{table}[h]
\centering\footnotesize
\setlength{\tabcolsep}{4pt}
\caption{\textbf{Waymo 2s under the protocol of the official DGGT code.} Frames: scored frames of each clip. DGGT as reported in its paper.}
\label{tab:supp_dggt_protocol}
\begin{tabular}{@{}llcc@{}}
\toprule
Method & Frames & PSNR$\uparrow$ & SSIM$\uparrow$ \\
\midrule
DGGT~\cite{ff_dggt} & see text & 27.41 & 0.846 \\
Ours & $0$--$15$ & 28.27 & 0.850 \\
Ours & intermediate only & 27.57 & 0.829 \\
\bottomrule
\end{tabular}
\end{table}

\section{Additional Visualizations}
\begin{figure*}[p]\centering
\includegraphics[width=.90\textwidth,height=.80\textheight,keepaspectratio]{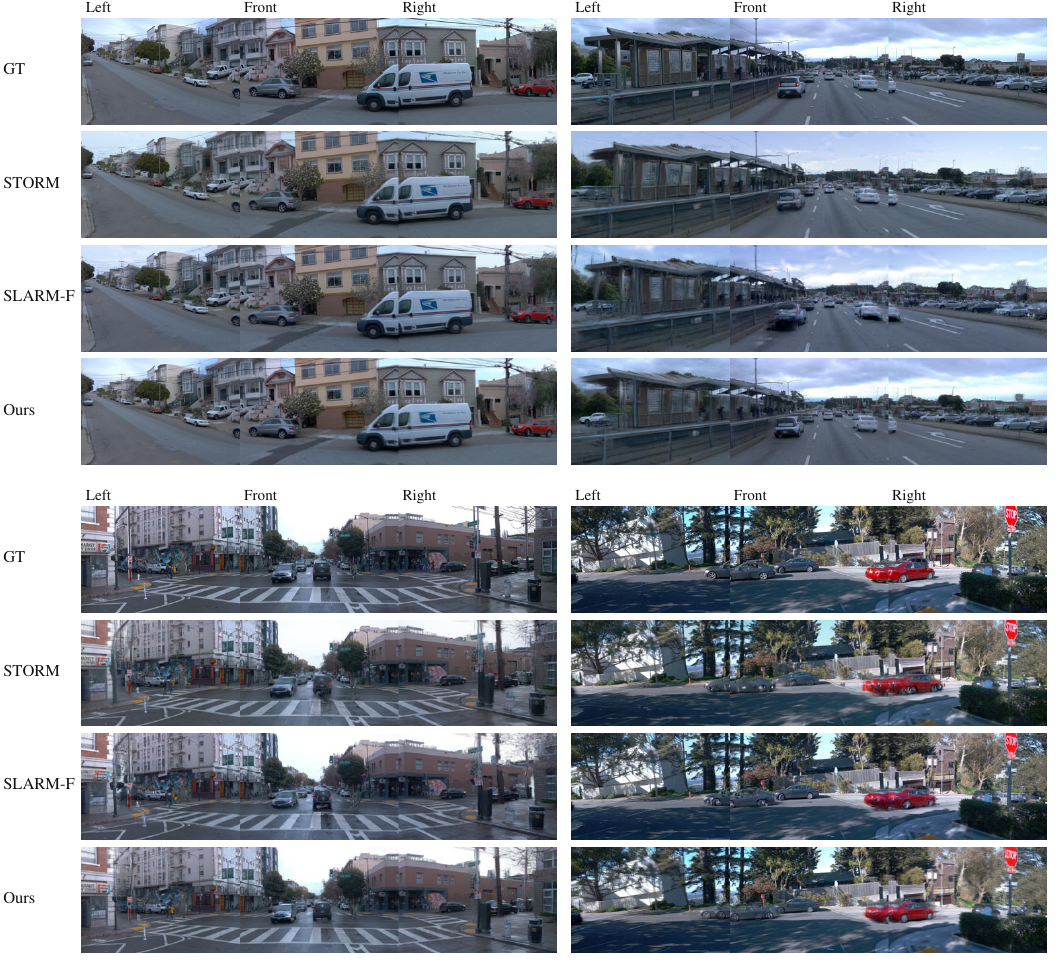}
\caption{\textbf{Additional Waymo comparisons.} Rows show the ground truth (GT), STORM, SLARM-F, and ours, illustrating nearby vehicles, occlusion, intersections, and lateral traffic.}\label{fig:supp_qual1}\end{figure*}

\begin{figure*}[p]\centering
\includegraphics[width=.90\textwidth,height=.80\textheight,keepaspectratio]{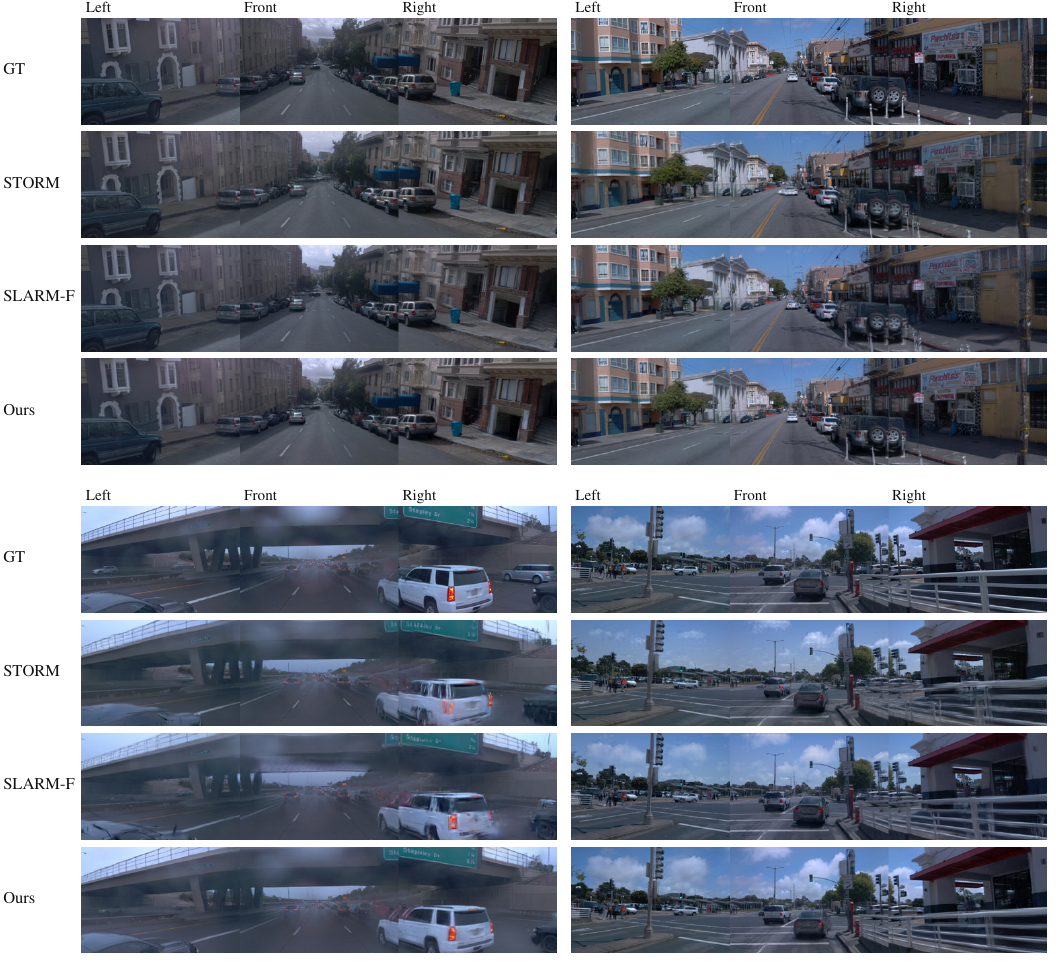}
\caption{\textbf{Additional Waymo comparisons.} Same layout as Fig.~\ref{fig:supp_qual1}, illustrating urban structure, road traffic, and complex visibility.}\label{fig:supp_qual2}\end{figure*}

\begin{figure*}[p]\centering
\includegraphics[width=\textwidth]{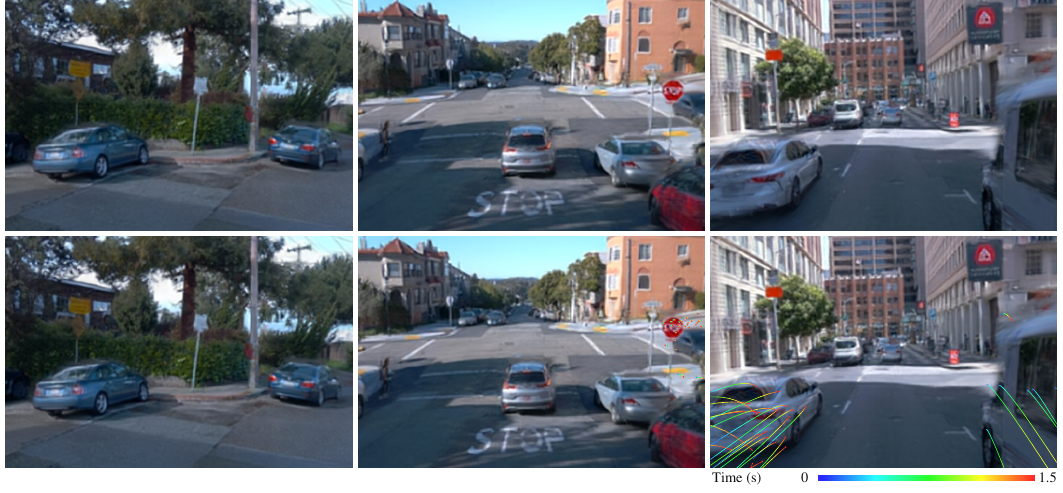}
\caption{\textbf{Additional novel-view renderings and Gaussian trajectories.} Renderings from a raised camera and predicted trajectories of Gaussians on moving vehicles, with the same 2s three-view model and 0--1.5\,s interval as Fig.~4 of the main paper.}\label{fig:supp_traj}\end{figure*}

\begin{figure*}[p]
\centering
\includegraphics[width=\textwidth]{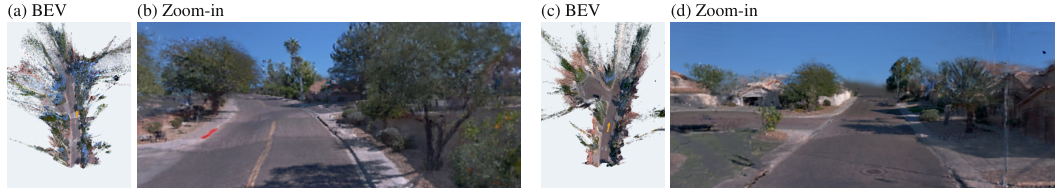}
\caption{\textbf{Scene-level static reconstruction and local views.} Each BEV and zoom-in pair shares a static Gaussian representation merged from approximately 20\,s of RGB. Orange arrows locate the virtual cameras.}
\label{fig:ff_bev}
\end{figure*}

\begin{figure*}[p]\centering
\includegraphics[width=\textwidth]{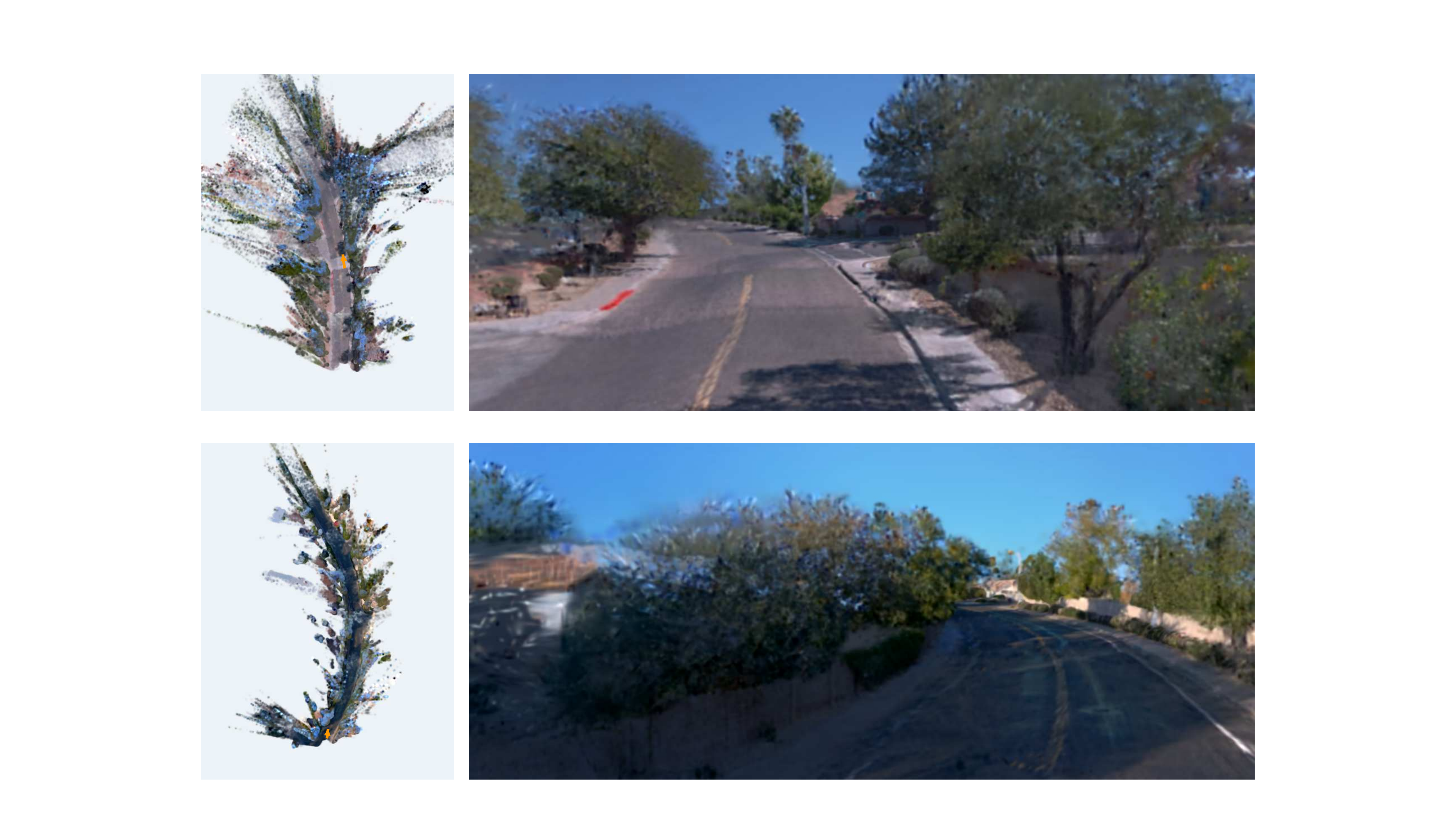}
\caption{\textbf{Wider static reconstructions.} Two driving scenes and zoom-ins, each merged from approximately 20\,s of static Gaussians. The top row is a wider view of the scene in Fig.~\ref{fig:ff_bev}(a,b).}\label{fig:supp_static1}\end{figure*}

\begin{figure*}[p]\centering
\includegraphics[width=\textwidth]{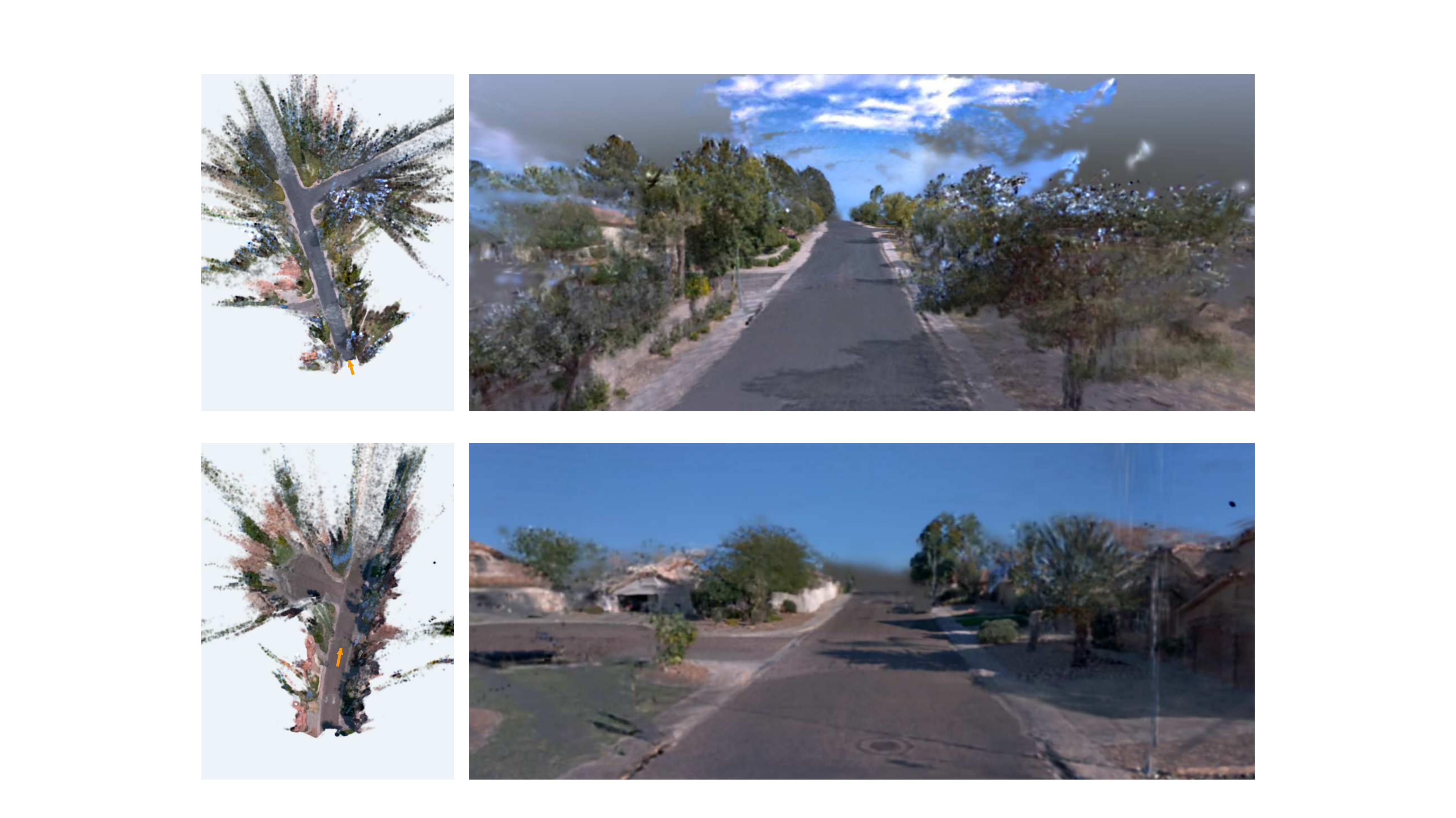}
\caption{\textbf{Additional static reconstructions.} Zoom-ins under the same conditions as Fig.~\ref{fig:supp_static1}. The bottom row is a wider view of the scene in Fig.~\ref{fig:ff_bev}(c,d).}\label{fig:supp_static2}\end{figure*}

Figs.~\ref{fig:supp_qual1} and~\ref{fig:supp_qual2} compare eight additional driving scenes with STORM and SLARM-F, Fig.~\ref{fig:supp_traj} shows predicted Gaussian trajectories, and Figs.~\ref{fig:ff_bev}--\ref{fig:supp_static2} show scene-level static reconstruction.

\paragraph{Scene-level static reconstruction.}
Our calibrated metric cameras place static Gaussians predicted by the 8s model over approximately 20\,s into a common coordinate system, without per-scene optimization. Fig.~\ref{fig:ff_bev} shows road layout in bird's-eye view and road markings, vegetation, and buildings in local views.

\clearpage